\documentclass{article}

\usepackage{microtype}
\usepackage{graphicx}
\usepackage{makecell}
\usepackage{subfigure}
\usepackage{booktabs} 

\usepackage{hyperref}

\usepackage[accepted]{icml2025}

\usepackage{amsmath}
\usepackage{amssymb}
\usepackage{mathtools}
\usepackage{amsthm}

\usepackage[capitalize,noabbrev]{cleveref}

\theoremstyle{plain}

\theoremstyle{definition}

\theoremstyle{remark}

\usepackage[textsize=tiny]{todonotes}

\icmltitlerunning{Submission and Formatting Instructions for ICML 2025}

\begin{document}

\definecolor{customcolor}{RGB}{240,230,207}

\twocolumn[
\icmltitle{\makecell{\textbf{IKUN}: \textbf{I}nitialization to \textbf{K}eep snn training and generalization \\ great with s\textbf{U}rrogate-stable varia\textbf{N}ce}}




\begin{icmlauthorlist}
\icmlauthor{Da Chang}{yyy,comp}
\icmlauthor{Deliang Wang}{sch}
\icmlauthor{Xiao Yang}{yyy,comp}
\end{icmlauthorlist}

\icmlaffiliation{yyy}{Shenzhen Institute of Advanced Technology, Chinese Academy of Sciences, Shenzhen,China}
\icmlaffiliation{comp}{Pengcheng Laboratory,Shenzhen,China}
\icmlaffiliation{sch}{Aerospace Information Research Institute, Chinese Academy of Sciences, Beijing, China}


\icmlcorrespondingauthor{Da Chang}{changda24@mails.ucas.ac.cn}

\icmlkeywords{Machine Learning, ICML}

\vskip 0.3in
]



\printAffiliationsAndNotice{}  

\begin{abstract}
Weight initialization significantly impacts the convergence and performance of neural networks. While traditional methods like Xavier and Kaiming initialization are widely used, they often fall short for spiking neural networks (SNNs), which have distinct requirements compared to artificial neural networks (ANNs).

To address this, we introduce \textbf{IKUN}, a variance-stabilizing initialization method integrated with surrogate gradient functions, specifically designed for SNNs. \textbf{IKUN} stabilizes signal propagation, accelerates convergence, and enhances generalization. Experiments show \textbf{IKUN} improves training efficiency by up to \textbf{50\%}, achieving \textbf{95\%} training accuracy and \textbf{91\%} generalization accuracy.

Hessian analysis reveals that \textbf{IKUN}-trained models converge to flatter minima, characterized by Hessian eigenvalues near zero on the positive side, promoting better generalization. The method is open-sourced for further exploration: \href{https://github.com/MaeChd/SurrogateVarStabe}{https://github.com/MaeChd/SurrogateVarStabe}.

\end{abstract}

\section{Introduction}
\label{introduction}
Spiking Neural Networks (SNNs) have garnered significant attention in both academia and industry in recent years \cite{ghosh2009spiking}\cite{bohte2000spikeprop}\cite{tonnelier2007event}\cite{gerum2021integration}. As a biologically-inspired computational model, SNNs emulate the spiking behavior of biological neurons, exhibiting unique advantages such as sparsity and event-driven computation. These characteristics endow SNNs with great potential in low-power and efficient computing, particularly in scenarios demanding high energy efficiency and real-time processing, such as neuromorphic hardware and edge computing. However, compared to traditional Artificial Neural Networks (ANNs), the discontinuity and complex temporal dynamics of SNNs pose significant challenges for training and optimization.

In neural network research, initialization methods are widely regarded as a critical factor influencing network performance and are key to successful training. Effective initialization contributes to network performance in the following three aspects:

\begin{itemize}
    \item Maintaining stable signal propagation across layers to prevent signal magnitude imbalance \cite{lecun2002efficient}\cite{glorot2010understanding}.
    \item Facilitating gradient flow to mitigate issues such as vanishing or exploding gradients during training \cite{he2015delving}\cite{he2016deep}.
    \item Accelerating convergence and improving training efficiency \cite{glorot2010understanding}.
\end{itemize}

However, the spiking nature of SNNs and the surrogate gradient (SG) mechanism adopted in some training methods introduce new challenges for initialization. The SG mechanism defines a smooth gradient approximation to address the non-differentiable nature of SNNs. However, recent studies indicate that this approach introduces bias: SG provides a biased gradient approximation even in deterministic networks with well-defined gradients. Moreover, in SNNs, the gradients derived from SG often do not align with the actual gradients of the surrogate loss, potentially affecting the effectiveness of training \cite{gygax2024elucidating}.

A more prominent issue lies in the inconsistency of SNN activation functions between forward and backward propagation. This raises a critical question:

\begin{center}
\makecell{
Can existing neural network initialization methods\\ be directly applied to SNNs?
}
\end{center}

Traditional ANN initialization methods typically assume continuous activation functions, overlooking the nonlinear, discrete spiking characteristics of SNNs. This assumption may not hold for SNNs. The dynamic behavior of spiking neurons fundamentally differs from the properties of continuous activation functions, making it difficult to maintain stable signal magnitudes in SNNs \cite{wu2018spatio}. Additionally, the impact of the SG mechanism on signal propagation in backpropagation is often neglected, exacerbating the problem. Existing ANN initialization methods are not optimized for SNN training based on SG, potentially limiting the generalization ability of SNNs, especially in deeper networks.

To address these issues, this paper proposes a novel initialization method specifically designed for SNNs, named \textbf{IKUN (\textbf{I}nitialization to \textbf{K}eep SNN training and generalization great with s\textbf{U}rrogate-stable varia\textbf{N}ce}). This method integrates the spiking characteristics of SNNs with the SG mechanism, optimizing both signal propagation and gradient dynamics. By controlling signal and gradient variances, IKUN achieves stability in both forward and backward propagation. Simultaneously, it leverages the properties of surrogate gradients to improve training efficiency and generalization performance.

The main contributions of this paper are as follows:
\begin{itemize}
    \item \textbf{Proposing the IKUN initialization method}: Through theoretical derivations and experimental validations, we design a dynamic initialization method that adapts to surrogate gradient characteristics, significantly enhancing the training and generalization performance of SNNs.
    \item \textbf{Plug-and-play compatibility}: The IKUN method is compatible with various surrogate gradient functions (e.g., sigmoid, tanh, and linear surrogate gradients), demonstrating strong versatility.
    \item \textbf{Experimental validation}: Experimental results on benchmark datasets show that compared to traditional initialization methods, IKUN achieves notable improvements in training stability, convergence speed, and generalization performance.
\end{itemize}

\section{Related Work}

\subsection{Weight Initialization}

Weight initialization is a critical component of deep neural network training. Its primary goal is to assign appropriate initial weights to the network at the start of training, ensuring that the variance of signals remains stable as they propagate through the network. This stability helps to effectively prevent issues such as gradient vanishing or gradient explosion. To address these challenges, researchers have proposed several classic weight initialization methods.

LeCun et al.\cite{lecun2002efficient} introduced one of the earliest initialization methods focusing on variance stability. Their method aimed to stabilize signal propagation by optimizing the weight distribution, specifically designed for Sigmoid and Tanh activation functions. The initialization formula is as follows:
\begin{equation}
    \sigma_w = \sqrt{\frac{1}{fan\_in}}
\end{equation}

Subsequently, Xavier initialization \cite{glorot2010understanding} advanced this idea by ensuring consistency in signal variance during both forward and backward propagation when using symmetric activation functions (e.g., Tanh, Sigmoid). This improvement enhanced network training performance. The weight initialization formula is:
\begin{equation}
    \sigma_w = \sqrt{\frac{2}{fan\_in + fan\_out}}
\end{equation}

Xavier initialization stands out for simultaneously considering the number of input and output nodes (fan-in and fan-out), making it suitable for bidirectional nonlinear activation functions and stabilizing signal variance during propagation.

Kaiming initialization \cite{he2015delving} extended this approach to scenarios involving asymmetric activation functions, particularly ReLU \cite{glorot2011deep} and its variants. This method, tailored to the unidirectional characteristics of ReLU activation functions, only considers the influence of fan-in, similar to LeCun initialization. The formula is:
\begin{equation}
    \sigma_w = \sqrt{\frac{2}{fan\_in}}
\end{equation}

This strategy adapts well to unidirectional nonlinear activation functions like ReLU but requires additional adjustments to account for their asymmetry, ensuring stable signal propagation.

The above weight initialization methods have been widely applied in ANNs, providing reliable stability for signal variance during propagation. However, these methods were primarily designed for ANN models optimized using conventional gradient-based approaches. In SNNs, the introduction of surrogate gradient mechanisms results in significant differences in signal propagation characteristics compared to ANNs. This can lead to signal attenuation or obstructed gradient flow in SNNs.

To address these issues, our method builds upon the core principles of the aforementioned initialization techniques while incorporating the unique characteristics of surrogate activation functions. We designed a weight initialization strategy better suited for SNNs to mitigate signal degradation and gradient blockage, thereby enhancing the training efficiency and performance of the network.

\subsection{SNN Training}
\subsubsection{Overview of SNNs}

Spiking Neural Networks (SNNs) are computational models inspired by biological nervous systems \cite{ghosh2009spiking}. Unlike traditional Artificial Neural Networks (ANNs), neurons in SNNs communicate through discrete spikes at specific time points. This means that the communication between neurons is based on the timing and frequency of spikes, exhibiting complex nonlinear dynamic behaviors \cite{bohte2000spikeprop}. Two commonly used neuron models in SNNs are the IF (Integrate-and-Fire) model \cite{tonnelier2007event} and the LIF (Leaky Integrate-and-Fire) model \cite{gerum2021integration}.

The IF model is the simplest spiking neuron model, assuming that the membrane potential $V(t)$ of a neuron accumulates without any leakage. The change in membrane potential follows the equation:
\begin{equation}
    V(t) = V(t - 1) + I(t)*\mathrm{\Delta}t/C
\end{equation}
where $V(t)$ is the membrane potential at time $t$, $I(t)$ is the input signal to the neuron at time $t$, $\mathrm{\Delta}t$ is the time step, and $C$ is the membrane capacitance. When the membrane potential reaches or exceeds the threshold $V_{th}$, the neuron emits a spike and resets the membrane potential to a resting value $V_{reset}$.

\begin{figure}[ht]
    \centering
    \centerline{\includegraphics[width=0.5\columnwidth]{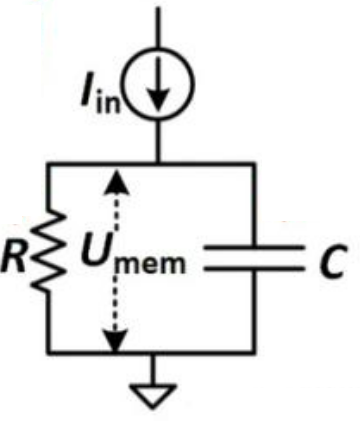}}
    \caption{Mechanism of the LIF neuron. This diagram illustrates the working principle of an LIF neuron, modeled as an RC circuit with leakage. When an input spike $I(t)$ causes the membrane potential $V(t)$ to accumulate and reach the threshold $V_{th}$, the neuron emits a spike and resets its potential to the resting value $V_{reset}$. If the threshold is not reached, the membrane potential gradually decays to the resting level, governed by the time constant $\tau = RC$.}
    \label{lif}
\end{figure}

The LIF model builds on the IF model by adding a leakage property. If the neuron does not receive sufficient stimuli to reach the threshold within a certain period, its membrane potential gradually decays back to the resting level $V_{reset}$. This model abstracts the neuron as an RC circuit, as shown in \cref{lif}, and follows the equation:
\begin{equation}
    \tau\frac{dV}{dt} = - V(t) + RI(t)
\end{equation}
where $\tau = RC$ is the time constant, $R$ is the membrane resistance, $C$ is the membrane capacitance, $I(t)$ is the input signal at time $t$, and $V(t)$ represents the membrane potential at time $t$. To integrate this neuron model into a deep learning framework, the equation can be discretized, yielding:
\begin{equation}
    V(t) = \left( {1 - \frac{1}{\tau}} \right)*V\left( {t - 1} \right) + \frac{1}{C}*I(t)
\end{equation}
From this equation, it is evident that the membrane potential at the current time depends on the potential at the previous time step, implying that the membrane potential accumulates over time until it reaches the threshold. Once the threshold is reached, the neuron emits a spike, and the membrane potential resets to the resting level or below.

\begin{figure}[ht]
    \centering
    \centerline{\includegraphics[width=\columnwidth]{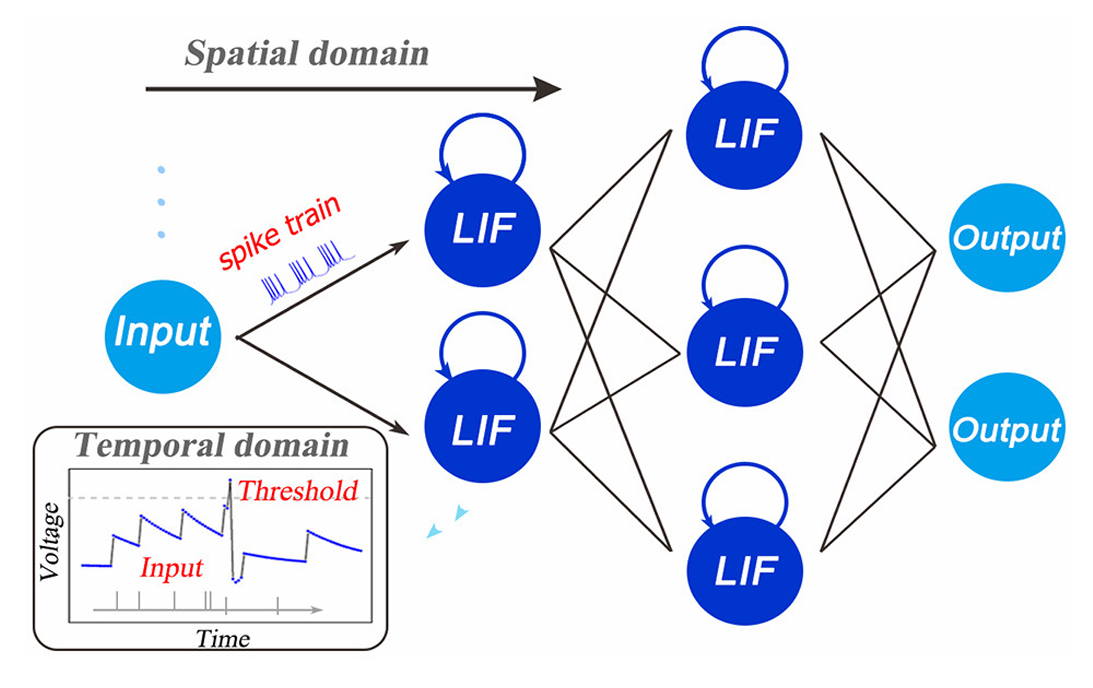}}

    \caption{Temporal logic in SNNs, adapted from \cite{wu2018spatio}. The figure illustrates the basic principles of temporal logic in SNNs. Multiple neurons are connected via weighted synapses to form a layered network structure. Input signals are encoded before being fed into the network, with common encoding methods including temporal encoding and rate encoding. Temporal encoding represents information through spike intervals, capturing the temporal properties of the original data, while rate encoding uses spiking frequency, where stronger stimuli correspond to higher spike rates. These properties make SNNs efficient for processing time-sensitive data.}
    \label{snn_temporal}
\end{figure}

With the basic neuron module, an SNN can be constructed by connecting neurons with weighted synapses, forming a hierarchical structure as shown in \cref{snn_temporal}. Since the input signals for SNNs must be discrete, the data fed into the network needs to be encoded. Two commonly used encoding methods are rate encoding and temporal encoding. Temporal encoding represents information through the time intervals of spikes, capturing important information about the temporal structure of the original data \cite{comsa2020temporal}. Rate encoding encodes information using the spiking frequency of neurons, where stronger stimuli are typically represented by higher spiking frequencies \cite{jin2018hybrid}.

Inspired by ANN models, SNNs have incorporated operations such as convolution and pooling, supporting multi-layer fully connected structures \cite{xu2018csnn}. Combining the feature extraction capabilities of convolutional neural networks with the biological plausibility of spiking neural networks, SNNs have evolved into deeper and more complex structures, offering outstanding processing speed and low energy consumption.

\subsubsection{Surrogate Gradient Mechanism in SNNs}

Currently, the training methods for SNNs can be categorized into three main approaches: training based on biological synaptic plasticity \cite{bi1998synaptic}, backpropagation algorithms using surrogate gradients \cite{neftci2019surrogate} \cite{ledinauskas2020training}, and training based on artificial neural network (ANN) conversion \cite{rueckauer2017conversion}. This paper primarily focuses on the weight initialization problem in SNNs trained using surrogate gradient-based backpropagation algorithms.

Due to the discrete and non-differentiable nature of SNN activation functions, traditional backpropagation algorithms cannot directly optimize their weights. To address this issue, surrogate gradient methods were introduced. These methods approximate the gradient of spiking activation functions with continuous and differentiable functions, enabling the application of backpropagation for optimization. Specifically, the core of surrogate gradient methods is to replace the non-differentiable spiking activation function's gradient with a continuous surrogate function (e.g., the approximation function for the IFNode \cite{fang2023spikingjelly}) for gradient computation.

It is worth noting that the challenge of non-differentiable activation functions is not unique to SNNs. In ANNs, a common method for gradient estimation is the Straight-Through Estimator (STE) \cite{bengio2013estimating}. For example, Yin et al. \cite{yin2019understanding} analyzed the training stability of models using STE and studied the impact of different types of STE on weight updates. Building on such studies, we hypothesize that the characteristics of surrogate gradients may significantly influence gradient flow in SNNs. It is well known that gradient explosion or vanishing in ANNs often results from inappropriate weight initialization or activation function selection. A similar mechanism might affect training stability in SNNs, albeit in a different form.

To better understand this issue, consider an extreme case: if the true gradient of the spiking activation function (e.g., the Dirac $\delta$ function) is directly used in backpropagation, its discontinuity and potentially large magnitude would lead to highly unstable training. Under such circumstances, no initialization strategy could ensure stable training. In contrast, the backpropagation process in ANNs relies on consistent activation functions during forward and backward propagation, and appropriate initialization strategies can effectively ensure stable gradient flow.

\begin{figure}[ht]
    \centering
    \centerline{\includegraphics[width=\columnwidth]{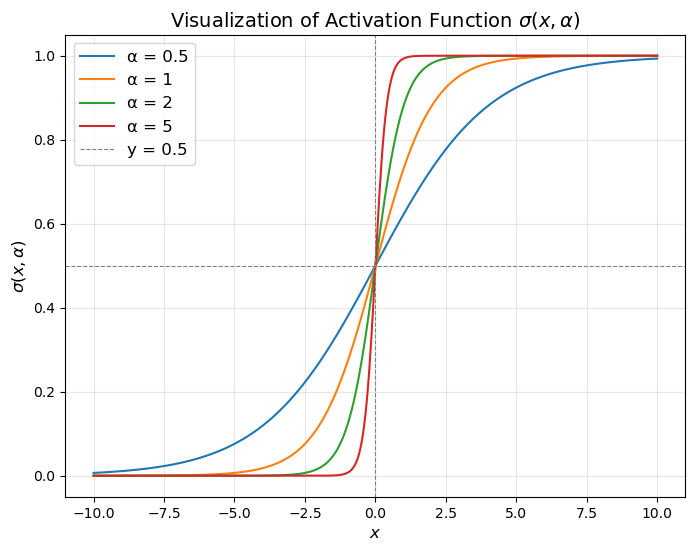}}
    \caption{Activation function curve of $\sigma(x, \alpha) = \frac{1}{1 + \exp(-\alpha x)}$. As $\alpha$ increases, the function gradually approaches a step function.}
    \label{snn_activation}
\end{figure}

SNNs fall between these two extremes. On one hand, their activation functions can be adjusted using surrogate gradients. The design of surrogate gradients allows for a trade-off between approximating the true spiking activation function and maintaining training stability. For instance, as shown in \cref{snn_activation}, for $\sigma(x, \alpha) = \frac{1}{1 + \exp(-\alpha x)}$, increasing the hyperparameter $\alpha$ makes the function increasingly resemble a step function. As the surrogate gradient becomes closer to the true spiking activation function, the optimization process tends to become less stable. On the other hand, current weight initialization schemes for SNNs typically reuse ANN methods or rely on default configurations, lacking optimizations tailored to SNN-specific characteristics. This raises questions about their applicability and effectiveness.

Therefore, the tunability of surrogate gradients provides a space for balancing approximation accuracy and training stability. At the same time, it highlights the need to consider the potential impact of SNN initialization schemes on gradient flow and model training. Exploring initialization strategies better suited for SNNs may be a key direction for improving their training performance.

\subsection{Analysis of Existing Methods for Improving SNN Training}

Existing research has made some progress in alleviating the challenges of SNN training, but exploration of weight initialization remains limited. Below are several typical approaches and their limitations:

\textbf{1. Regularization and Normalization Techniques} \\
Regularization methods (e.g., Dropout \cite{srivastava2014dropout}) have limited effectiveness in SNNs because the sparsity of spiking signals inherently provides a regularization effect. Direct application of these techniques may disrupt the temporal information of the signals. Similarly, normalization methods (e.g., Batch Normalization \cite{ioffe2015batch} and Layer Normalization \cite{lei2016layer}) struggle to mitigate the gradient flow issues caused by surrogate gradients, contributing little to improving training performance.

\textbf{2. Dynamic Adjustment Methods} \\
Strategies that dynamically adjust surrogate gradient parameters, such as $\alpha$, can achieve rapid learning in the early stages of training and stable convergence in later stages. However, these methods are highly dependent on parameter tuning and primarily focus on dynamic optimization during training, without directly addressing initial issues in signal propagation and gradient distribution.

\textbf{3. Initialization Methods} \\
Weight initialization studies often adopt strategies designed for artificial neural networks, such as Xavier \cite{glorot2010understanding} and Kaiming \cite{he2015delving}, without fully considering the spiking activation mechanisms and surrogate gradient characteristics of SNNs. Traditional methods may lead to issues such as:
\begin{itemize}
    \item Imbalanced signal propagation variance, limiting the expressive capacity of deeper networks.
    \item Misaligned scaling of surrogate gradients, adversely affecting optimization effectiveness and convergence speed.
\end{itemize}

Although existing methods have partially alleviated the challenges of SNN training, research on weight initialization requires further investigation. Designing initialization methods tailored to the unique characteristics of SNNs holds the potential to optimize signal propagation and gradient dynamics, thereby significantly improving training efficiency and generalization performance.

\section{Method}
\subsection{Core Idea of the IKUN Initialization Method}

\paragraph{Variance-Stable Initialization}
Variance-stable initialization aims to ensure that the variance of signals remains constant as they propagate between network layers, thereby avoiding issues like gradient vanishing or explosion, especially in deep networks. This is crucial for accelerating convergence and improving model performance \cite{lecun2002efficient}\cite{glorot2010understanding}\cite{he2015delving}. On one hand, the spiking mechanism of neurons in the forward propagation of SNNs differs from traditional activation functions (such as ReLU), altering the statistical properties of signal propagation. On the other hand, in backpropagation, smooth surrogate functions (e.g., $\text{ATan}(x)$) are used to compute gradients to avoid instability caused by directly handling derivatives of impulse functions. This inconsistency between forward and backward processes requires a comprehensive consideration of the variance effects of both mechanisms to achieve balanced propagation of signals and gradients. Therefore, a targeted initialization strategy needs to be designed.

\paragraph{Core Idea}
In forward propagation, maintain the variance of each layer's output consistent with its input, and in backpropagation, preserve stable gradient variance to address the differences arising from the unique discrete spiking activation mechanisms of SNNs and surrogate gradient methods. The key to variance-stable initialization is to adapt to these characteristics, ensuring efficient signal propagation in deep networks and enhancing training stability and model performance.

\subsection{Theoretical Guarantee of the IKUN Initialization Method}
\subsubsection*{Theorem 1: Surrogate-Stable Variance Initialization (IKUN)}

In SNNs, suppose the membrane potential update formula is
\begin{equation}
H[t] = \tau V[t-1] + \sum_{i=1}^{m} w_i X_i[t],
\end{equation}

The weight initialization strategy satisfies $w_i \sim \mathcal{N}(0, \sigma_W^2)$, and the threshold is set as $V_{\text{threshold}} = \mu_H$. By the following weight variance condition:
\begin{equation}
\label{eq:IKUNv1} 
\sigma_W^2 = \frac{\alpha}{\mathrm{fan}_{\mathrm{in}} \cdot \sigma_X^2 \cdot \mathbb{E}[f'(H[t])^2]},
\end{equation}

we can ensure variance stability of signals during forward propagation while avoiding gradient vanishing or explosion during backpropagation. Here, $\alpha$ is a given hyperparameter.

We provide a detailed derivation in the proof section of the appendix. 

\paragraph{Remarks}

Usually, we assume the input follows a normal distribution with mean 0 and variance 1. Therefore, the simplified form of \cref{eq:IKUNv1} resembles Kaiming initialization, which we refer to as IKUN v1. Furthermore, we propose an improved initialization scheme:
\begin{equation}
\sigma_W^2 = \frac{\alpha}{(\mathrm{fan}_{\mathrm{in}} + \mathrm{fan}_{\mathrm{out}}) \cdot \sigma_X^2 \cdot \mathbb{E}[f'(H[t])^2]},
\end{equation}

which we call IKUN v2 initialization, to more comprehensively balance the influence of input and output neurons.


\section{Experiments}
\subsection{Experimental Setup}

\textbf{Dataset} \\
This experiment evaluates the proposed method on the FashionMNIST dataset \cite{xiao2017fashion}. FashionMNIST is a classical image classification dataset consisting of 28×28 grayscale images across 10 categories. The training set contains 60,000 images, and the test set contains 10,000 images.

\begin{figure}[ht]
    \centering
    \centerline{\includegraphics[width=\columnwidth]{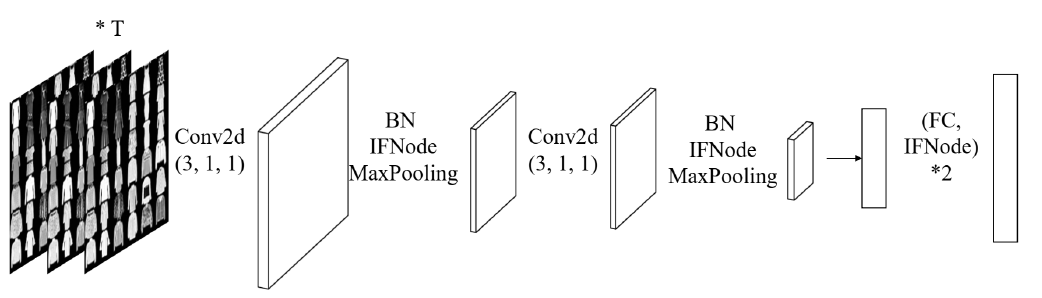}}
    \caption{Network architecture. The diagram illustrates the two-layer convolutional SNN architecture, where each layer comprises convolution, pooling, and spiking activation functions for efficient extraction and processing of temporal features in input data.}
    \label{framework}
\end{figure}

\textbf{Model Architecture} \\
The experiment uses a typical SNN model architecture—a \textbf{two-layer convolutional SNN}, which consists of two convolutional layers. Each layer includes convolution, pooling, and spiking activation functions. The specific structure is shown in \cref{framework}.

\textbf{Comparison Methods} \\
To validate the effectiveness of the proposed method, the following initialization methods are selected for comparison: Xavier Initialization \cite{glorot2010understanding}, Kaiming Initialization \cite{he2015delving}, LeCun Initialization \cite{bi1998synaptic}, and the default random initialization.

\textbf{Optimizers} \\
Two optimizers are used for training: SGD \cite{robbins1951stochastic} and Adam \cite{kingma2014adam}.

\textbf{Experimental Environment} \\
All experiments are conducted on a server equipped with an NVIDIA 1650Ti GPU with 4 GB of memory. The software environment is as follows: Operating System: Windows 10; Python version: 3.8; Deep Learning Framework: PyTorch 1.12.0; SNN Library: SpikingJelly 0.0.0.14; CUDA version: 11.3. These settings ensure the reliability, fairness, and reproducibility of the experiments.

\subsection{Evaluation Metrics}

\textbf{Loss Function} \\
The training loss is measured using Mean Squared Error (MSE), defined as:
\begin{equation}
\mathcal{L} = \frac{1}{N} \sum_{i=1}^N |y_i - \hat{y}_i|^2,
\end{equation}
where $N$ is the number of samples, and $y_i$ and $\hat{y}_i$ represent the ground truth and predicted values, respectively. The convergence efficiency and training stability are assessed by analyzing the rate of decline and fluctuation of the loss curve.

\textbf{Classification Accuracy} \\
Classification accuracy evaluates the generalization ability of the model on the test set and is defined as:
\begin{equation}
\text{Accuracy} = \frac{1}{M} \sum_{i=1}^M \mathbb{1}(\hat{y}_i = y_i),
\end{equation}
where $M$ is the number of test samples, and $\mathbb{1}(\cdot)$ is the indicator function, which equals $1$ if the prediction is correct and $0$ otherwise.

\textbf{Hessian Analysis} \\
To investigate training dynamics and optimization performance, we analyze the Hessian matrix characteristics of the model after training:

\begin{itemize}
    \item \textbf{Hessian Spectrum Distribution:} By analyzing the eigenvalues $\{\lambda_1, \lambda_2, \dots, \lambda_P\}$ of the Hessian matrix, we assess the flatness of the loss landscape. A smaller maximum eigenvalue or narrower spectrum distribution is often associated with better generalization performance \cite{wang2018identifying}.
    
    \item \textbf{Hessian Matrix Trace:} The trace is defined as the sum of diagonal elements:
    \begin{equation}
    \text{Tr}(H) = \sum_{i=1}^P H_{ii}.
    \end{equation}
    Lower $\text{Tr}(H)$ values indicate that the model resides in a flatter region at the end of optimization, typically corresponding to better generalization ability \cite{yao2020pyhessian}.
\end{itemize}

By combining these analyses, we provide a comprehensive evaluation of training dynamics and local curvature, offering a solid basis for interpreting model performance and comparing initialization methods.

\subsection{Experimental Results}

\begin{figure}[ht]
    \centering
    \begin{tabular}{@{}c@{\hspace{1cm}}c@{}}
        
        \includegraphics[width=0.45\columnwidth]{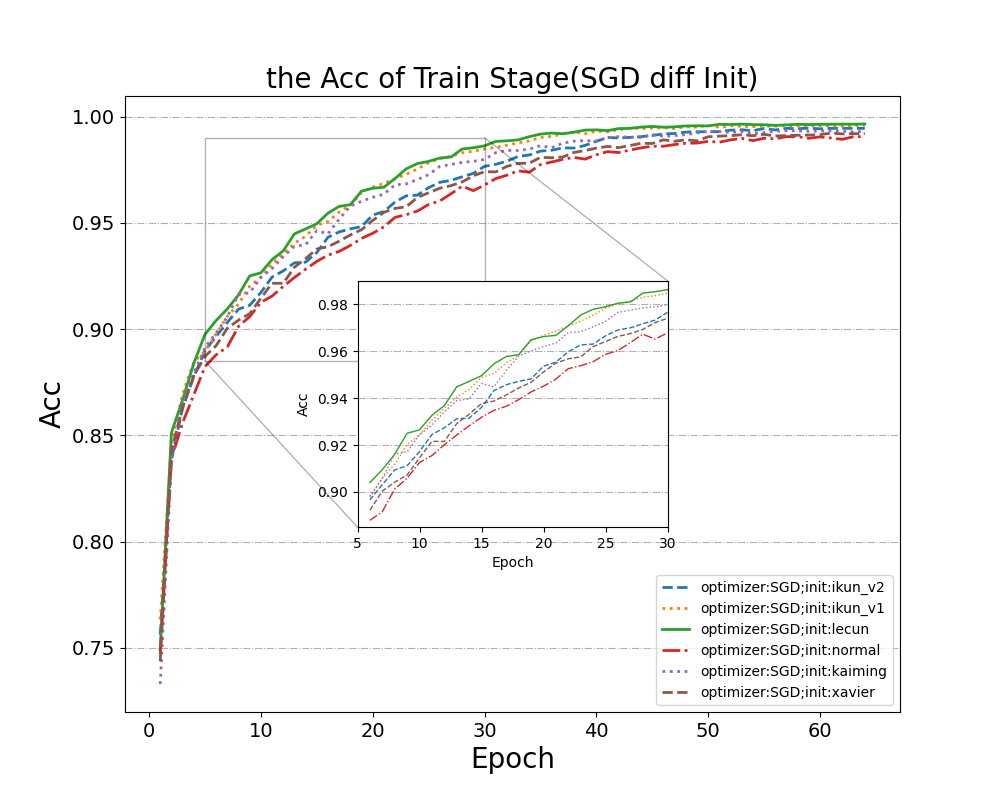} &
        \includegraphics[width=0.45\columnwidth]{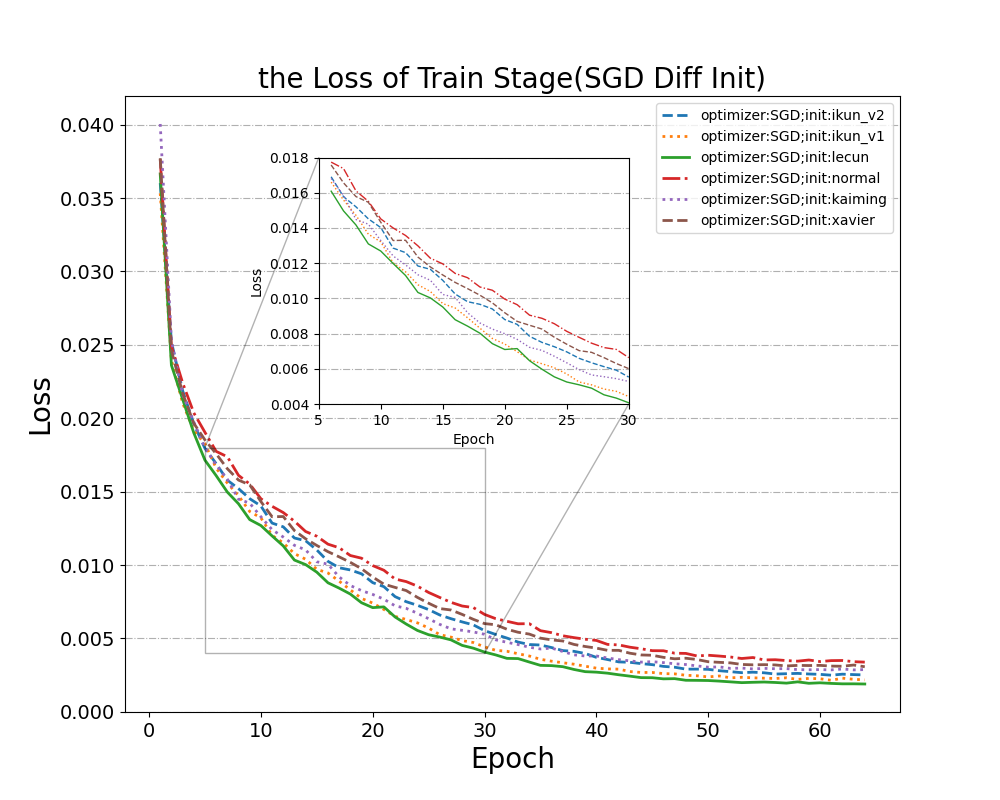} \\
        (a) SGD Train Accuracy & (b) SGD Train Loss \\
        \includegraphics[width=0.45\columnwidth]{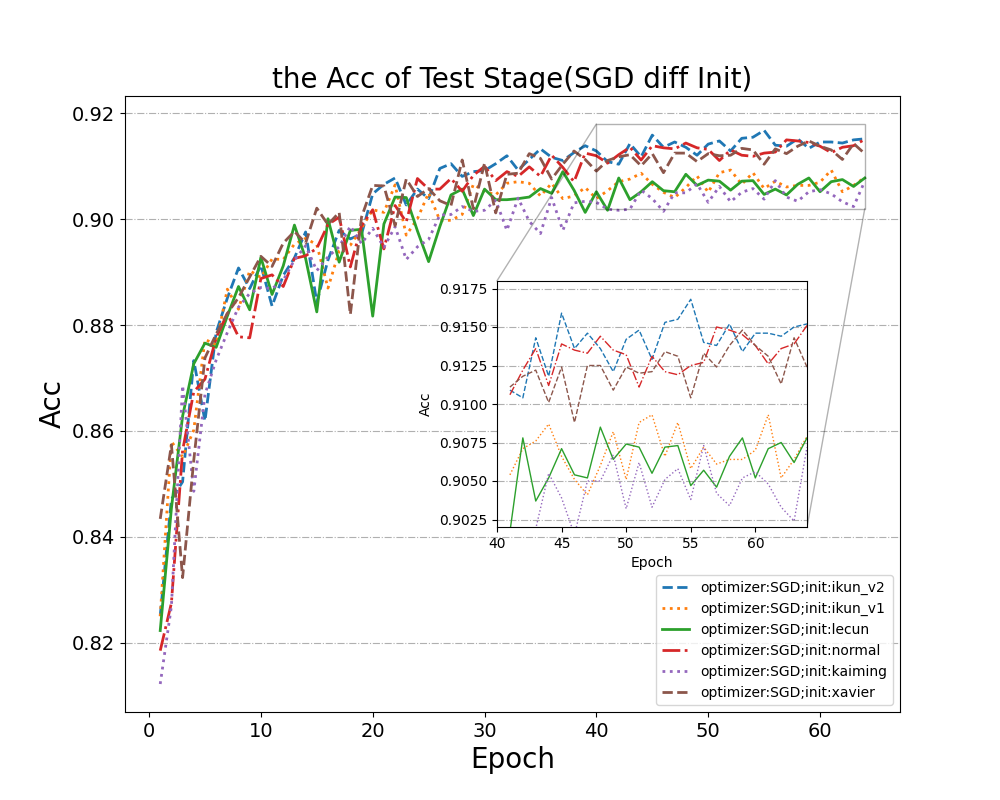} &
        \includegraphics[width=0.45\columnwidth]{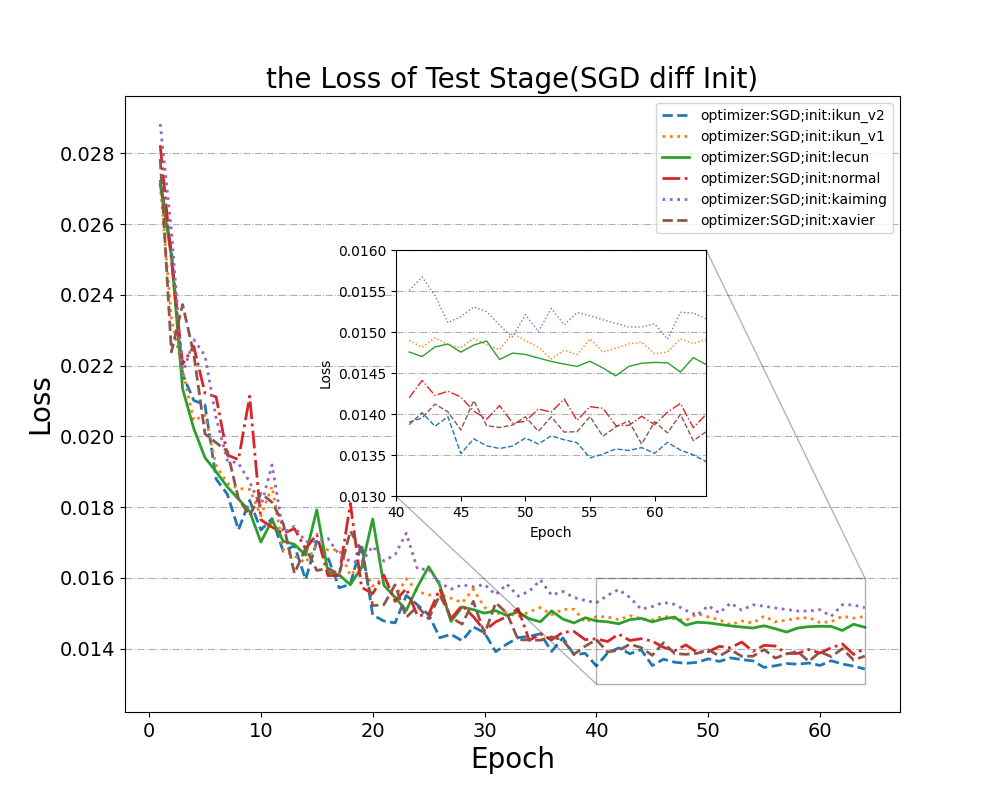} \\
        (c) SGD Test Accuracy & (d) SGD Test Loss
    \end{tabular}
    \caption{(a) and (c) show the changes in training and testing accuracy under the SGD optimizer, while (b) and (d) depict the training and testing loss curves. IKUN initialization achieves faster loss reduction in the early training stages and maintains higher stability, but in certain cases, its performance is slightly worse than other initialization methods.}
    \label{table:sgd_comparison}
\end{figure}

\begin{figure}[ht]
    \centering
    \begin{tabular}{@{}c@{\hspace{1cm}}c@{}}
        \includegraphics[width=0.45\columnwidth]{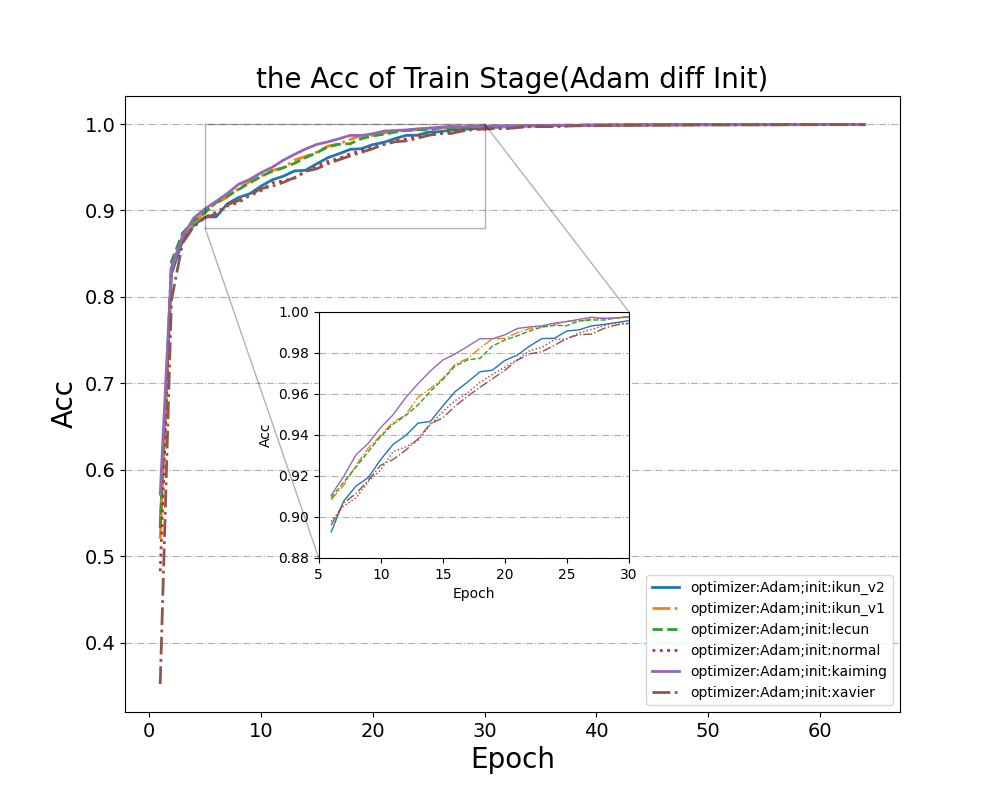} &
        \includegraphics[width=0.45\columnwidth]{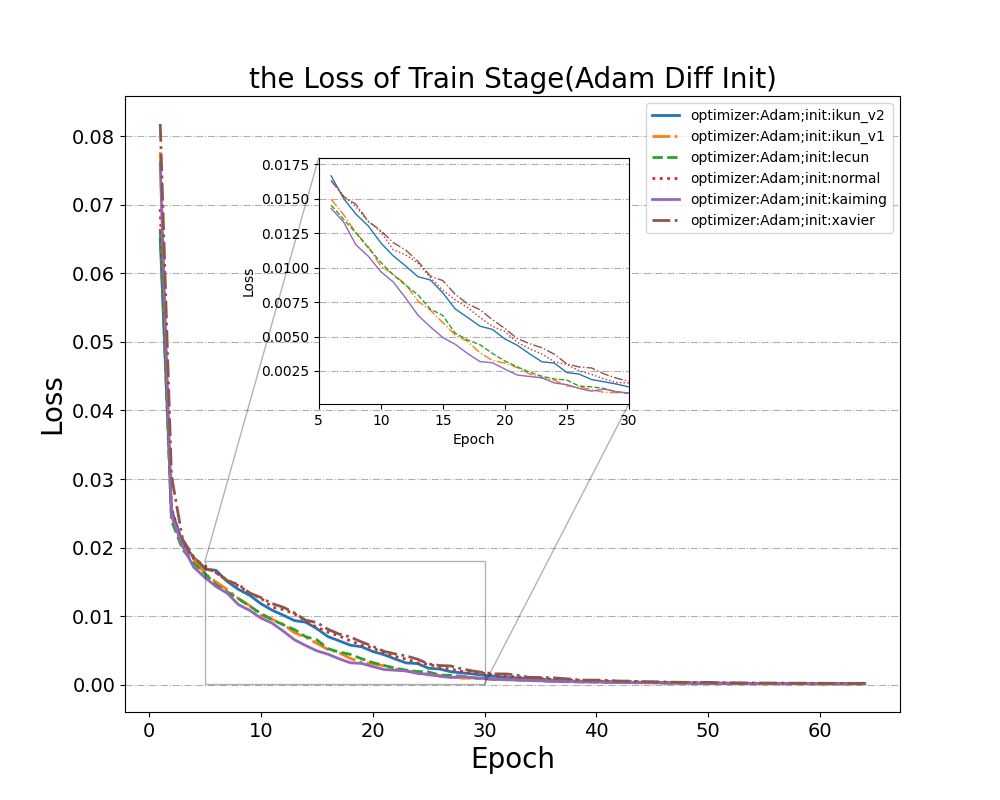} \\
        (a) Adam Train Accuracy & (b) Adam Train Loss \\
        \includegraphics[width=0.45\columnwidth]{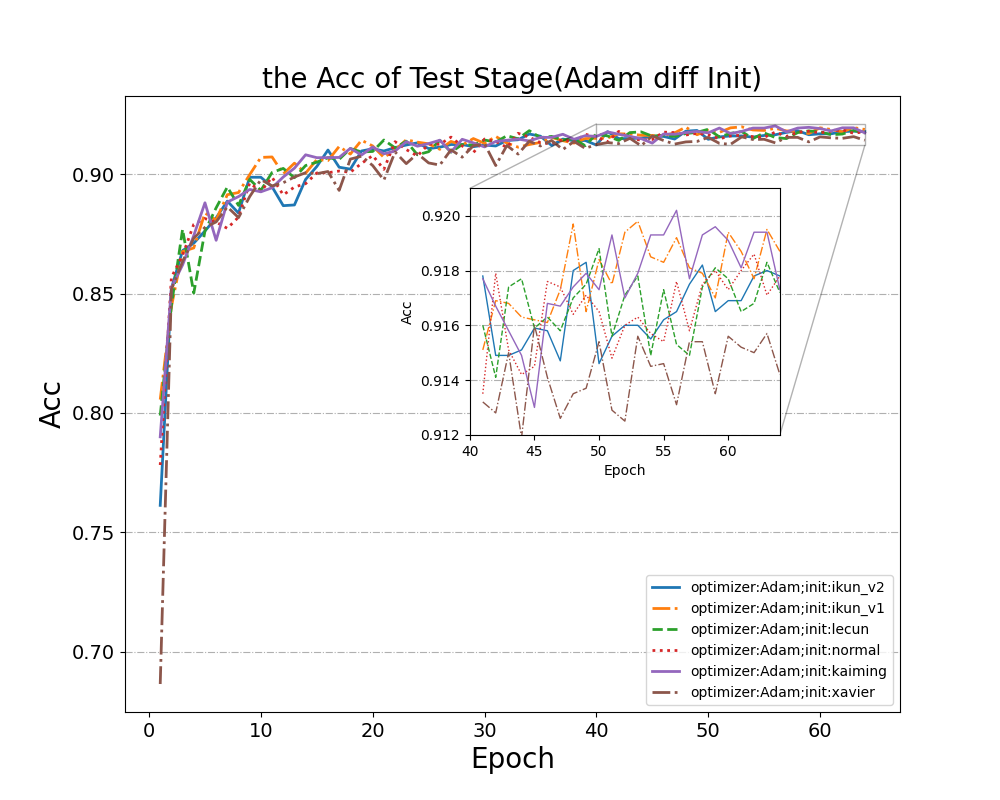} &
        \includegraphics[width=0.45\columnwidth]{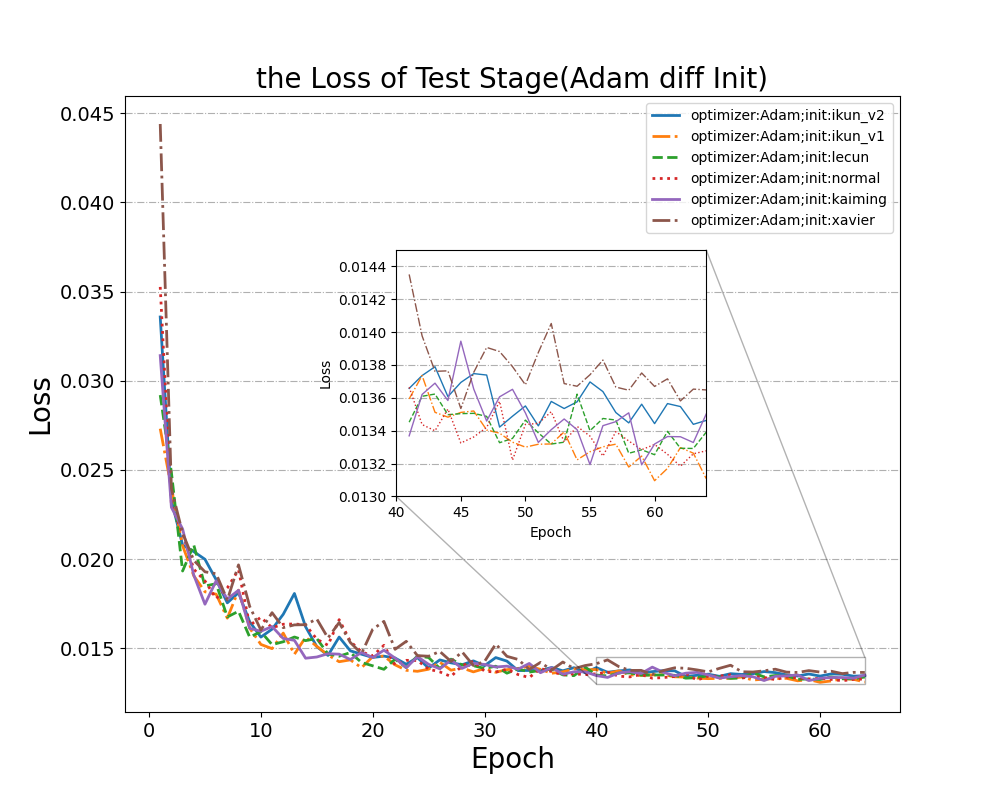} \\
        (c) Adam Test Accuracy & (d) Adam Test Loss
    \end{tabular}
    \caption{(a) and (c) show the training and testing accuracy curves under the Adam optimizer, while (b) and (d) illustrate the loss curves. Although IKUN initialization achieves faster loss reduction in the early stages, it underperforms compared to Kaiming and LeCun initialization in certain scenarios.}
    \label{table:Adam_comparison}
\end{figure}

We present the training and testing accuracy, as well as the training and testing loss curves, under both Adam and SGD optimizers. From \cref{table:sgd_comparison} and \cref{table:Adam_comparison}, it is evident that the IKUN initialization method demonstrates faster loss reduction in the early stages of training compared to other initialization methods, while maintaining better stability in later stages. However, the results also show that IKUN initialization does not always exhibit significant advantages. For instance, under the Adam optimizer, IKUN initialization performs worse than Kaiming initialization; under the SGD optimizer, its performance is slightly inferior to LeCun initialization. It is worth noting that traditional initialization methods fail to achieve optimal performance simultaneously on both optimizers, highlighting a potential advantage of our method.

Additionally, the differences in testing errors during the later stages of training are relatively minor, making it difficult to fully demonstrate the effectiveness of the method based solely on these results. To further analyze and compare the key metrics of the model during training, we adopt the following two schemes:

\textbf{Scheme 1:} Identify the training epochs when both training accuracy and testing accuracy reach predefined thresholds (e.g., 90\% or 95\%). This scheme evaluates the impact of different initialization methods on training speed and generalization ability, focusing on the model's performance during the transition from training to generalization.

\textbf{Scheme 2:} Identify the training epochs when testing accuracy reaches its peak and record the corresponding Hessian Trace (reflecting the curvature characteristics of the parameter space). This scheme emphasizes evaluating the stability and robustness of the model at its best performance point, providing a comprehensive reference for assessing initialization methods.

The specific experimental results are shown in \cref{compare1} and \cref{compare2}:

\cref{compare1} shows that under different training thresholds (we selected training accuracy greater than 95\% and testing accuracy greater than 91\%), the distribution of training epochs varies significantly. By introducing our IKUN initialization method, training efficiency improves substantially: training epochs are reduced by up to \textbf{59.38\%} under the SGD optimizer and \textbf{42.31\%} under the Adam optimizer. This demonstrates that IKUN initialization effectively shortens training time while maintaining high-quality training performance.

\cref{compare2} reveals the relationship between the best testing accuracy, the corresponding testing loss, and the Hessian Trace. Lower testing loss and smaller absolute Hessian Trace values are typically associated with better generalization performance, indicating that the model parameters are optimized in a smoother and more stable space. The results show that IKUN initialization achieves superior Hessian Trace values under both SGD and Adam optimizers (\textbf{5.27} and \textbf{1.17}, respectively). Traditional methods often achieve optimal Hessian Trace values under either SGD or Adam alone, whereas IKUN initialization performs well under both optimizers.

\begin{table*}[t]
    \begin{center}
    \resizebox{1.2\columnwidth}{!}{
    \begin{tabular}{lcccccc}
    \toprule
         Init & \makecell{SGD \\Epoch} & \makecell{Adam\\ Epoch} & \makecell{SGD \\Train Acc} & \makecell{SGD \\Test Acc}  & \makecell{Adam \\Train Acc} & \makecell{Adam \\Test Acc} \\
    \midrule
        \texttt{\textbf{IKUN v2}}  & \textbf{26} & \textbf{15} & 97.00 &  91.05 & 96.10 &  91.01\\
        \texttt{IKUN v1}  &  - & 16  & - &  - & 97.72 &  91.16\\
        \texttt{LeCun}  & - & 17 & - &  - & 97.72 &  91.07\\
        \texttt{Normal}  & 29 & 21 & 96.78 &  91.00 & 98.10 &  91.36 \\
        \texttt{Kaiming}  & - & 17  & - &  - & 98.68 &  91.08 \\
        \texttt{Xavier}  & 27  & 26  & 96.93 &  91.12 & 98.90 &  91.02\\
        \hline
    \end{tabular}}
    \end{center}
    \caption{Results for Scheme 1. This table compares the training epochs required to achieve specific accuracy thresholds on training and testing sets for different initialization methods. IKUN v2 significantly reduces training epochs under both SGD and Adam optimizers, by 59.38\% and 42.31\%, respectively, demonstrating advantages in training speed and generalization ability. "—" indicates that the result did not meet the specified conditions before training ended.}
    \label{compare1}
\end{table*}

\begin{table*}[t]
    \begin{center}
    \resizebox{1.2\columnwidth}{!}{
    \begin{tabular}{lcccc}
    \toprule
         Init & \makecell{SGD \\Test Acc} & \makecell{Adam \\Test Acc}  &  \makecell{SGD \\Hessian Trace} & \makecell{Adam \\Hessian Trace} \\  
    \midrule
        \texttt{\textbf{IKUN v2}}  & 91.68 & 91.83 & \textbf{5.27} & \textbf{1.17} \\
        \texttt{IKUN v1}  & 90.93 & 91.98 & 29.39 & 8.21 \\
        \texttt{LeCun}  & 90.90 & 91.88  & 3.26  &  -205.18\\
        \texttt{Normal}  & 91.51 & 91.86 & 52.70 & -53.59\\
        \texttt{Kaiming}  & 90.73 & 92.02 & 5.44 & 3.74\\
        \texttt{Xavier}  & 91.48 & 91.48 & 161.99 & -728.51 \\
        \hline
    \end{tabular}}
    \end{center}
    \caption{Results for Scheme 2. This table compares the testing loss and Hessian Trace values at the best testing accuracy for different initialization methods. IKUN v2 achieves lower Hessian Trace values under both SGD and Adam optimizers (5.27 and 1.17, respectively), demonstrating smoother and more stable parameter spaces and superior generalization performance compared to other methods.}
    \label{compare2}
\end{table*}

In summary, these results indicate that the proposed initialization method not only significantly improves training efficiency but also achieves better training behavior and generalization ability across different optimizer settings, offering new perspectives and methodological support for SNN model optimization.

\begin{figure}[ht]
    \centering
    \begin{tabular}{@{}c@{\hspace{1cm}}c@{}}
        \includegraphics[width=0.4\columnwidth]{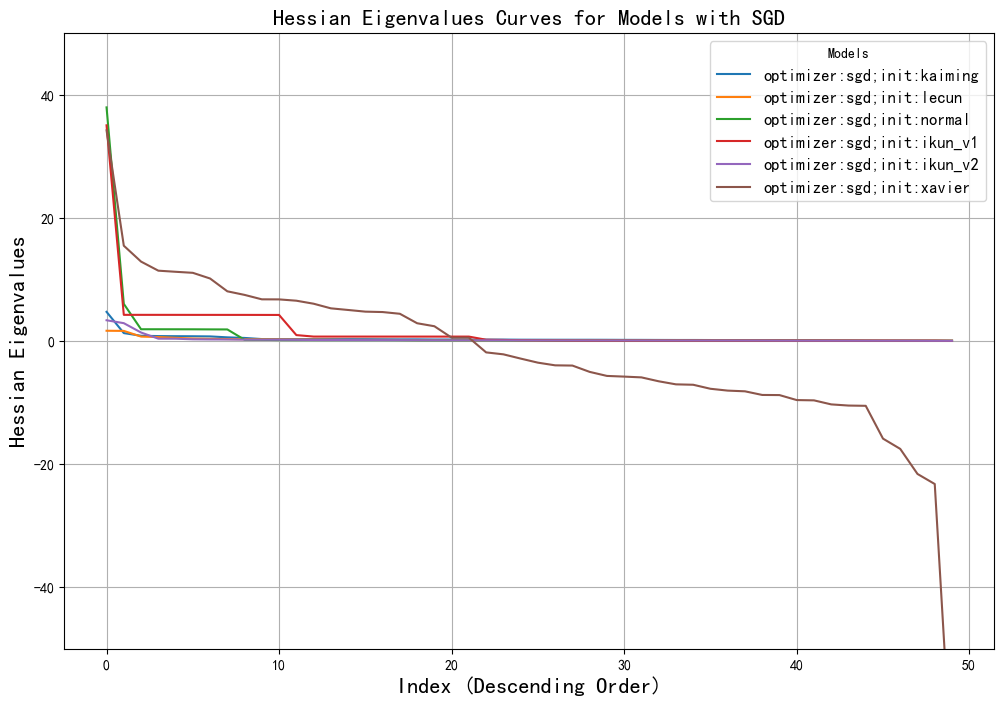} &
        \includegraphics[width=0.4\columnwidth]{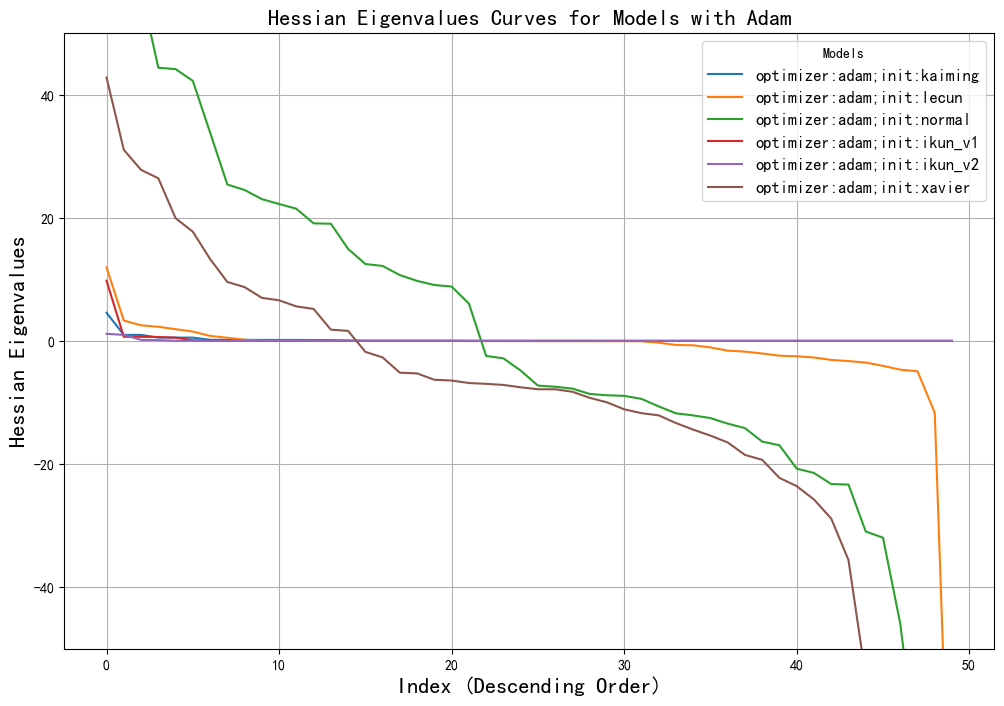} \\
        (a)& (b) \\
    \end{tabular}
    \caption{(a) shows the top 50 ranked Hessian eigenvalues using the SGD optimizer. Both IKUN and Kaiming initialization yield eigenvalues close to 0 and positive, indicating a flat optimization region with good curvature characteristics, better generalization, and robustness. (b) shows the top 50 ranked Hessian eigenvalues using the Adam optimizer, further validating the generalization performance of IKUN initialization in flat minima regions under different optimizer settings.}
    \label{hessain_rank}
\end{figure}

\subsection{Performance Validation}

In the experimental results section, we validated the effectiveness of the IKUN initialization method in terms of training efficiency and testing performance. To further explore its inherent advantages, we used Hessian eigenvalue distribution analysis as a tool to deeply investigate the optimization characteristics of the method. The distribution of Hessian eigenvalues reflects the curvature properties of the loss function's local region, serving as an important indicator for evaluating the generalization ability and optimization stability of the model \cite{wang2018identifying}.

We used the PyHessian package \cite{yao2020pyhessian} to compute and analyze the Hessian eigenvalues of the neural network, focusing on the performance comparison between the IKUN initialization method and other mainstream initialization methods.

\subsubsection{Comparison of Hessian Eigenvalue Distributions}

\begin{figure}[ht]
    \centering
    \includegraphics[width=\columnwidth]{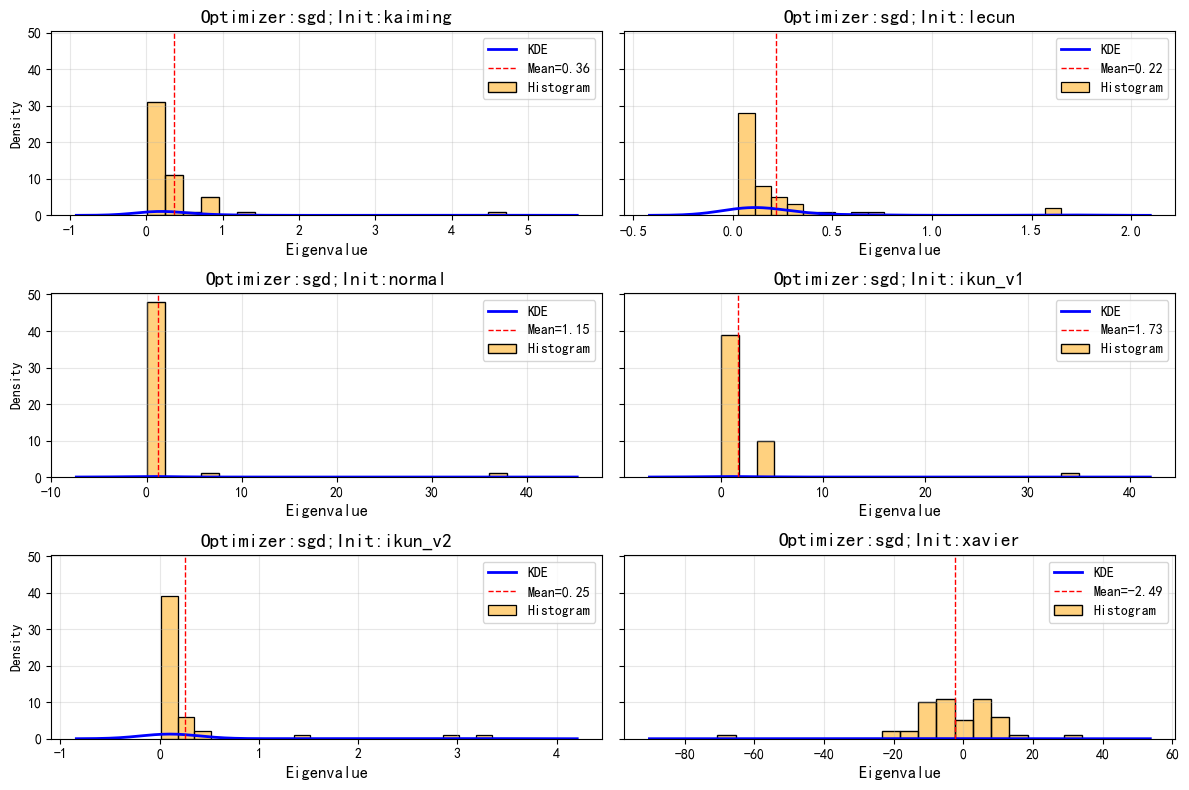}
    \caption{Hessian eigenvalue density distribution with the SGD optimizer. The figure compares the density distribution of Hessian eigenvalues for different initialization methods under the SGD optimizer. IKUN and Kaiming initialization exhibit concentrated distributions close to 0, while traditional methods (e.g., LeCun, Xavier, and Normal) display broader distributions, indicating less stable optimization curvature.}
    \label{hessain_density_sgd}
\end{figure}

\begin{figure}[ht]
    \centering
    \includegraphics[width=\columnwidth]{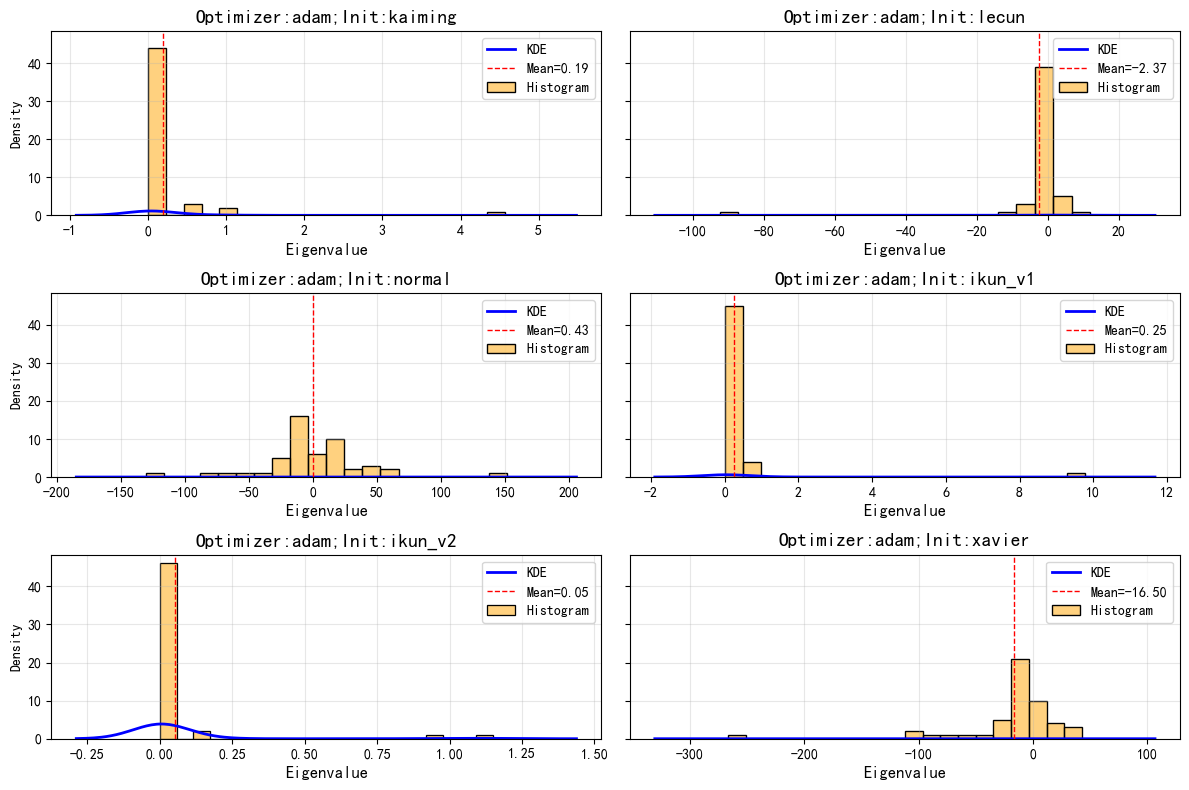}
    \caption{Hessian eigenvalue density distribution with the Adam optimizer. The figure shows the density distribution of Hessian eigenvalues under the Adam optimizer. The IKUN initialization method demonstrates a compact distribution of positive values, further supporting its superior optimization stability and generalization performance compared to traditional methods.}
    \label{hessain_density_adam}
\end{figure}

\textbf{IKUN vs. Kaiming Initialization: Eigenvalues Close to 0 and Positive} \\
As shown in \cref{hessain_rank}, \cref{hessain_density_sgd}, and \cref{hessain_density_adam}, the Hessian eigenvalues of IKUN and Kaiming initialization are almost entirely positive and close to 0, indicating that the optimization ends in flat minima regions. These curvature characteristics are typically associated with better generalization ability and enhanced robustness to parameter perturbations. Although Kaiming initialization, as the default method for CSNNs, performs well, it is not optimal; in contrast, IKUN initialization further improves generalization performance.

\textbf{Other Methods: Wider Eigenvalue Distribution with Positive and Negative Values} \\
Traditional methods (e.g., LeCun, Xavier, and Normal initialization) exhibit a broader Hessian eigenvalue distribution, including both positive and negative values. This indicates convexity in certain directions and concavity in others, suggesting that these methods may remain stuck in saddle-point regions, reducing optimization efficiency and parameter robustness. Furthermore, while these methods show strong exploratory behavior during training, their performance on the test set is often slightly inferior to that of the IKUN method.

\subsubsection{Comprehensive Analysis}

Based on the experimental results and the Hessian analysis in this section, the main advantages of the IKUN initialization method can be summarized as follows:

\textbf{1. Improved Generalization Ability:} The model tends to converge to flat minima regions with low curvature, resulting in greater robustness and reduced sensitivity to unseen data, thereby enhancing generalization performance.

\textbf{2. Stable Optimization Process:} Positive eigenvalues prevent saddle-point issues, leading to higher optimization efficiency.

\section{Conclusion}

We proposed the IKUN initialization method tailored for Spiking Neural Networks (SNNs). By leveraging the characteristics of surrogate gradients, this method significantly improves training stability and model generalization performance. Both theoretical analysis and experimental validation demonstrate the superior performance of the IKUN method: the Hessian eigenvalue distribution is almost entirely centered around zero, indicating that model parameters reside in flat minima regions—a property closely associated with better generalization ability. Compared to traditional methods (which exhibit both positive and negative Hessian eigenvalues), the IKUN method greatly reduces the impact of negative eigenvalues, avoiding stagnation at saddle points or regions with abrupt curvature changes. Furthermore, the potential of combining this method with other optimization strategies (e.g., dynamic learning rates, enhanced regularization) warrants further investigation.

Due to time and computational resource constraints, our experiments were conducted on a two-layer convolutional SNN using a fixed subset of the Fashion-MNIST dataset. Despite these limitations, the results indicate that our method demonstrates significant advantages in Hessian eigenvalue distribution, generalization performance, and test set outcomes.

Future research will focus on validating the effectiveness of the proposed method on larger datasets (such as CIFAR-10 and ImageNet) and more complex model architectures. The preliminary findings of this study provide a new perspective on SNN training optimization and generalization performance research. Future work will aim to further validate its potential value through broader experimentation.

\nocite{langley00}

\bibliography{ref}
\bibliographystyle{icml2025}




\end{document}